\documentclass[journal]{IEEEtran}
\usepackage{amsmath,amssymb}
\usepackage{booktabs,array,graphicx,url}
\usepackage[hidelinks]{hyperref}
\title{Observation Quality in DSM-Assisted Aerial Localization:\\
A Benchmark and Base-Anchored Uncertainty Calibration}
\author{Pengyu~Yin, Jie~Pan, Junxiang~Tan, Mengqi~Fu, Kun~Cheng, and~Yifan~Liu%
\thanks{This work was supported by the Aerospace Information Research Institute,
Chinese Academy of Sciences, under Grant E6E3190301.}%
\thanks{This work has been submitted to the IEEE for possible publication.
Copyright may be transferred without notice, after which this version may no
longer be accessible.}%
\thanks{\textit{(Corresponding author: Junxiang Tan.)}}%
\thanks{Pengyu Yin, Jie Pan, Junxiang Tan, and Mengqi Fu are with the Aerospace
Information Research Institute, Chinese Academy of Sciences, Beijing 100094,
China (e-mail: yinpy@aircas.ac.cn; panjie@aircas.ac.cn; tanjx@aircas.ac.cn;
1043879125@qq.com).}%
\thanks{Kun Cheng and Yifan Liu are with the School of Electronic Engineering,
Beijing University of Posts and Telecommunications, Beijing 100876, China
(e-mail: kcheng@bupt.edu.cn; yifanliu@bupt.edu.cn).}}
\hypersetup{pdfauthor={Pengyu Yin, Jie Pan, Junxiang Tan, Mengqi Fu, Kun Cheng, Yifan Liu}}
\begin{document}
\raggedbottom
\maketitle
\begin{abstract}
Aerial map matching provides position observations for navigation, but a
position error alone does not determine how an observation should be weighted.
We define five benchmark tasks: position availability
and agreement, grouped error structure, predictive uncertainty, controlled
updates, and temporal accumulation. We use digital-surface-model-assisted
localization to test whether base-anchored uncertainty calibration improves
the weighting of unchanged position estimates.
The method combines propagated image and map sensitivities with the second
moment of a compact configuration subset relative to the reported position.
Scale is fitted using separate position references without local surface-height
ground truth. Thirteen
covariance models are evaluated on two held-out query sets from locations
represented in the calibration data. Using reference residuals, we compare the
method with a constant full-matrix baseline. With all localization failures
included, the method reduces the normalized position-update risk, averaged equally
across locations, by 7.70\% on 728 conditions and 5.46\% on a separate 142
conditions under an independent prior. On the primary set, the equal-location
average negative log likelihood on reference residuals improves by 2.571 nat. At these
seen locations, exploratory base-only quality calibration and conditional-addition
controls leave the incremental update benefit of configuration disagreement
unresolved. Navigation benefit and real
temporal accumulation remain untested.
\end{abstract}

\begin{figure*}[t]
\centering
\includegraphics[width=\textwidth]{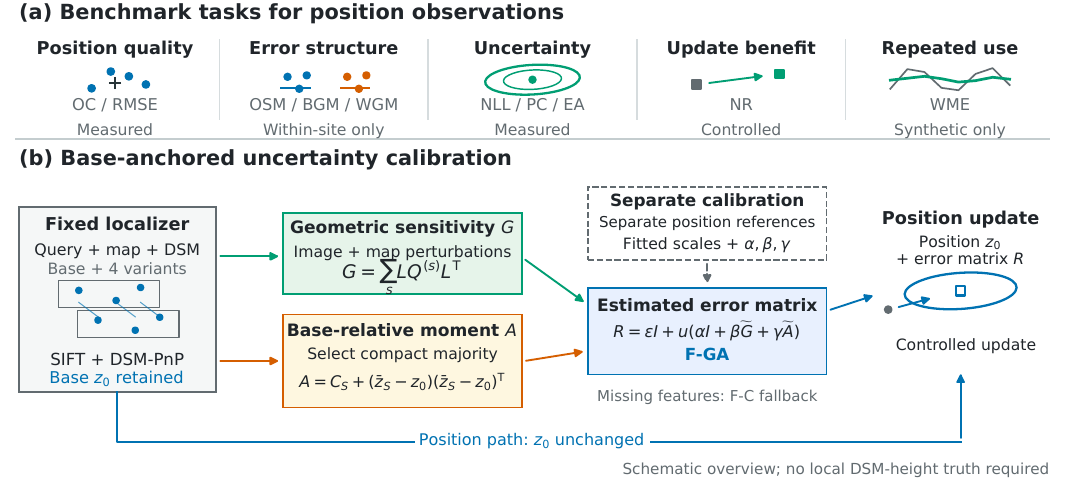}
\caption{Benchmark and method overview; all small plots are schematic.
(a) Five tasks characterize position observations and their use. (b) Geometric sensitivity
$G$ and base-relative moment $A$ predict error matrix $R$ while preserving
position $z_0$. Solid arrows carry inference data; the dashed arrow supplies
parameters estimated from separate calibration data and fixed during evaluation.
Task labels indicate which tasks are evaluated on real data.
F-C is the constant full-matrix baseline; F-GA combines geometry and
anchoring. Real-data evaluation excludes temporal accumulation and
cross-frame group moments.}
\label{fig:anchored-overview}
\end{figure*}

\section{Introduction}
\label{sec:introduction}

A position estimated by aerial image-to-map matching can correct navigation
drift. Assigning too much weight to an unreliable observation can degrade the
estimate, whereas assigning too little weight to an accurate observation
reduces the benefit of the correction. A digital surface model (DSM)
supplies localization geometry, with uncertainty in both matching and the map.
Average position error alone misses availability, persistent errors and whether
reported uncertainty leads to useful updates.

In our development experiments, a geometry-and-configuration model improved
predictive likelihood relative to a constant full-matrix baseline but increased
controlled update risk under an independent prior. In some cases, it reduced
the weight of an already accurate base estimate. This mismatch illustrates why
uncertainty should be evaluated through the updates it induces, not only
through its predictive density.

Uncertainty-aware pose estimation incorporates image and three-dimensional
feature uncertainty~\cite{vakhitov2021uncertainty}. Post-hoc regression
calibration adjusts predictive uncertainty without changing the predicted
mean~\cite{levi2020evaluating,kuppers2022parametric}.
We ask whether geometric sensitivities and configuration disagreement can
improve the weighting of an unchanged position after calibration.
Configuration disagreement can be large because a single variant is unstable,
even when the base estimate is accurate.

We define a benchmark with five tasks:
availability and position agreement, grouped error structure, predictive
uncertainty, controlled position updates, and temporal accumulation.
Each task specifies the intended use, metric, evaluation sample and
interpretation. Controlled counterexamples show what a single error score
misses. The benchmark applies independently of the calibration method below.

We evaluate base-anchored calibration, which combines
propagated matching and map sensitivities with the displacement of a compact
configuration subset from the reported position, retaining internal spread and
base offset. Reference positions from a separate calibration set determine
the scale without local
DSM-height ground truth.
Figure~\ref{fig:anchored-overview} connects the benchmark tasks to this
fixed-position uncertainty pipeline and separates fitting from inference.

Thirteen covariance models use the same positions and failures on 728
validation and 142 inPlace conditions, isolating the effect of uncertainty
estimation. We compare complete calibration procedures against the constant
full-matrix baseline, F-C, and compare F-GA with its single-component controls
to assess feature contributions. A subsequent base-only quality model tests
whether the extra configuration runs add useful information.
The experiments use paired local map crops and previously represented
locations. They assess controlled updates rather than global retrieval or
navigation over a flight sequence.

\section{Related Work}
\label{sec:related-work}

Learned matchers such as SuperPoint--LightGlue, RoMa, and GIM-trained models
estimate image correspondences
\cite{detone2018superpoint,lindenberger2023lightglue,edstedt2024roma,shen2024gim}.
Our study addresses the subsequent estimation of horizontal position
uncertainty while holding a SIFT--DSM--PnP localizer fixed. This design
isolates uncertainty estimation rather than comparing matching systems.
OrthoLoC's paired UAV--orthographic setting separates localization from
retrieval~\cite{dhaouadi2025ortholoc}, whereas AnyVisLoc examines a broader
retrieval--matching--DSM--pose pipeline~\cite{ye2026anyvisloc}. We use the
paired setting, so the results concern localization within a supplied map region.

Uncertainty-aware pose estimation incorporates image and three-dimensional
feature uncertainty into residual covariances used for pose estimation and
refinement~\cite{vakhitov2021uncertainty}. Local least-squares covariance
estimates depend on the measurement model and Jacobian
rank~\cite{cerescovariance}. We construct geometric
sensitivities whose scales are calibrated from position references.
Post-hoc regression calibration includes fixed-mean scale
adjustment~\cite{levi2020evaluating} and multivariate covariance
calibration~\cite{kuppers2022parametric}. Building on these methods, we test
whether geometry and configuration features improve on constant isotropic
and full-matrix baselines, and how they affect position-update risk.

RIC-Loc combines reference-induced pose dispersion with propagated geometric
covariance for failure ranking and uses robust consensus to estimate
position~\cite{kang2026ricloc}. We instead estimate the error moment of an
unchanged base observation. Robust location and scatter
estimation~\cite{hubert2018mcd}, including Minimum-Diameter Averaging's
fixed-size subset selection~\cite{elmhamdi2020mda}, provide established tools
for summarizing hypotheses. We use that selection objective to construct a
base-relative feature; its gradient-aggregation guarantees do not extend to
this localization setting.

The benchmark organizes RMSE, moment and coverage diagnostics, together with
logarithmic scoring~\cite{gneiting2007proper}, around observation availability
and downstream use. Availability accounts for failures; coverage, area and direction assess
the reported density; update and temporal tasks reveal consequences hidden
by aggregate position error. Together these tasks distinguish uncertainty calibration,
position accuracy and navigation performance.

\begin{figure*}[t]
\centering
\includegraphics[width=\textwidth]{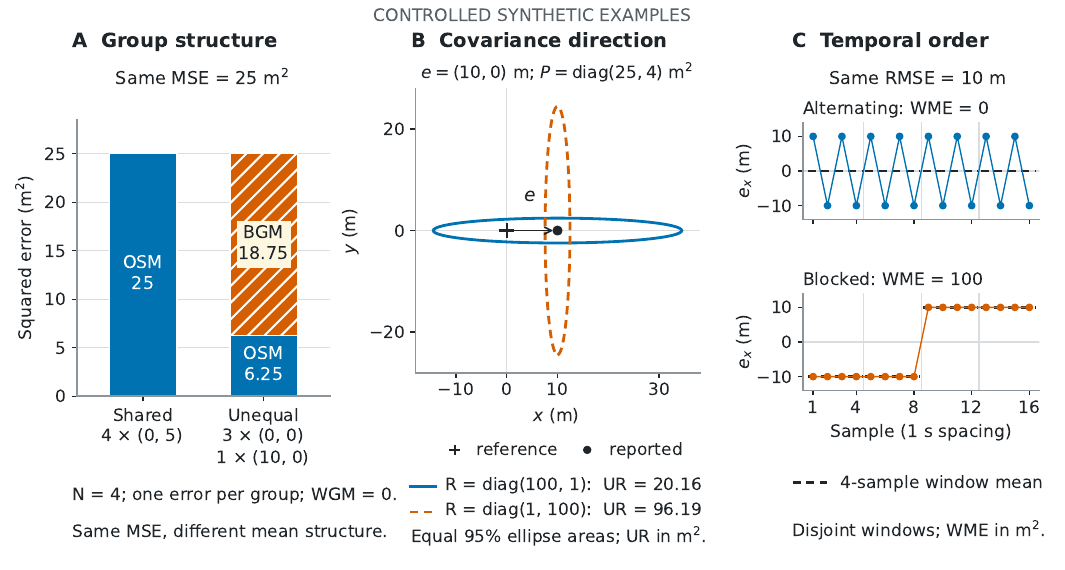}
\caption{Synthetic counterexamples with known reference positions.
(a) Equal sample count and mean squared error (MSE), different group structure; each observation
is a group. (b) Fixed residual, prior and ellipse area, different update
error after rotating the ellipse. (c) Reordering the same errors preserves
single-frame root mean squared error (RMSE) but changes window-mean error.}
\label{fig:compact-counterexamples}
\end{figure*}

\section{Benchmark Definition}
\label{sec:benchmark}

The benchmark evaluates position observations through five questions:
availability and accuracy, error structure across groups, uncertainty
calibration, the effect of a controlled update, and error accumulation over time.
Controlled counterexamples in Fig.~\ref{fig:compact-counterexamples} illustrate
why these properties can differ despite identical aggregate position error.
Tasks and metrics are specified before comparing methods;
Table~\ref{tab:metric-compact} lists the abbreviations.
The evaluation set contains $N$ predefined query conditions, termed
opportunities. Each completed attempt is classified as successful, failed or
rejected; unknown outcomes and unexecuted cases are recorded separately.
A successful output is a finite position. Each comparison specifies the
grouping, calibration/evaluation split, method versions, computational budget,
reference positions and covariance source (native, post-hoc or synthetic).
Calibration and evaluation sets are disjoint at the query level, with related
observations of each query assigned to the same set. Observations used during
method development are identified separately. Comparisons of uncertainty
models hold the position estimator and its computational budget fixed.

\begin{table*}[t]
\centering\small
\caption{Metric glossary. Units refer to per-observation values or the stated
aggregate. MV and SM are vector and matrix diagnostics, not scalar rankings.
The five benchmark tasks specify evaluation samples, failure treatment and aggregation.}
\label{tab:metric-compact}
\begin{tabular}{@{}p{0.26\textwidth}p{0.55\textwidth}p{0.13\textwidth}@{}}
\toprule
Labels & Meaning & Unit \\
\midrule
EC, OC, CA & Execution coverage, output coverage, conditional availability & 1 \\
MD, MedD, P95 & Mean, median, and 95th-percentile distance & m \\
$\mathrm{TR}_t$, $A_5$, $L_{1000}$ & Tail rate, 5 m reference-agreement rate, capped mean loss & 1, 1, m \\
MSE, RMSE & Mean squared error and its square root & m$^2$, m \\
OSM, BGM, WGM & Overall squared mean, between-group and within-group moments & m$^2$ \\
$\mathrm{RMSE}_{\mu}$, $\mathrm{RMSE}_{\rm mac}$ & Micro and macro RMSE & m \\
MV, SM & Mean residual vector, residual second-moment matrix & m, m$^2$ \\
Q, NLL & Squared normalized residual, negative log-likelihood & 1, nat \\
$\mathrm{PC}_{\alpha}$, $\mathrm{EA}_{\alpha}$, $\mathrm{DR}_v$ & Predictive coverage, ellipse area, directional ratio & 1, m$^2$, 1 \\
$\mathrm{UR}(P)$, $\mathrm{NR}(P)$ & Update risk and normalized update risk & m$^2$, 1 \\
$\mathrm{WME}(T)$, $\mathrm{EL}_t$, $\mathrm{ECnt}_t$ & Window-mean squared error, error-event length and count & m$^2$, sample, count \\
\bottomrule
\end{tabular}
\end{table*}

\subsection{Availability and Position Agreement}
\label{sub:q1-position}

Availability complements position accuracy: small errors on successful
queries alone do not describe performance on difficult queries that fail.
A prediction $z_i$ and reference $p_i^{\rm ref}$ must refer to the same physical
point and use compatible metric coordinates. In particular, camera centers and
ground query centers must be distinguished. Dataset-local horizontal $X,Y$ axes retain their local
designation unless a transformation to east/north or ENU is verified. Define
\begin{equation}
 d_i=z_i-p_i^{\rm ref},\qquad a_i=\|d_i\|.
 \label{eq:difference}
\end{equation}
If reference error is $u_i=p_i^{\rm ref}-p_i^{\rm true}$, then $d_i=e_i-u_i$,
where $e_i$ is true position error. Thus reference agreement does not certify
accuracy relative to the true position, particularly when reference
uncertainty is unquantified.

For $N_{\rm exec}$ terminal entries and $N_{\rm out}$ outputs,
$\mathrm{EC}=N_{\rm exec}/N$, $\mathrm{OC}=N_{\rm out}/N$, and
$\mathrm{CA}=N_{\rm out}/N_{\rm exec}$. Distance summaries use only the
compatible finite-output set $\mathcal V$, of size $n$: MD, MedD, P95, RMSE,
and $\mathrm{TR}_t=n^{-1}\sum_{\mathcal V}\mathbf1\{a_i>t\}$.
The fixed opportunity-level scores retain missing-output penalties:
\begin{equation}
 A_5=\frac{\sum_i\mathbf1\{i\in\mathcal V,\ a_i\leq5\,\mathrm m\}}{N},
 \qquad L_{1000}=\frac1N\sum_i\ell_i,
 \label{eq:legacy}
\end{equation}
where $\ell_i=\min(a_i,1000\,\mathrm m)$ for an output and
$1000\,\mathrm m$ for an explicit failure or rejection. These scores require
complete execution and compatible references; unknown outcomes cannot be
counted as failures. A fully executed all-failure method therefore has $A_5=0$
and $L_{1000}=1000\,\mathrm m$, while its conditional distance summaries
are undefined, as are summaries on any empty evaluation subset. The thresholds
are descriptive conventions, not safety limits.

\subsection{Grouped Error Structure}
\label{sub:q2-groups}

A repeatable offset requires different treatment from scattered errors of the
same average size. Groups, such as flights or map conditions, are specified before scoring.
For each nonempty group, let $b_g$ be its mean residual,
$S_g=n_g^{-1}\sum_{i\in g}(d_i-b_g)(d_i-b_g)^{\mathsf T}$,
$w_g=n_g/n$, and $b=\sum_gw_gb_g$. Then
\begin{equation}
 \underbrace{\frac1n\sum_{\mathcal V}\|d_i\|^2}_{\mathrm{MSE}}
 =\underbrace{\|b\|^2}_{\mathrm{OSM}}
 +\underbrace{\sum_gw_g\|b_g-b\|^2}_{\mathrm{BGM}}
 +\underbrace{\sum_gw_g\operatorname{tr}S_g}_{\mathrm{WGM}}.
 \label{eq:moments}
\end{equation}
Opposite offsets can cancel in the mean vector, but not in squared error.
The main pooled summaries are micro-weighted:
$\mathrm{RMSE}_{\mu}=\sqrt{\sum_gw_g\mathrm{MSE}_g}$.
The companion macro summary is
$\mathrm{RMSE}_{\rm mac}=\sqrt{G^{-1}\sum_g\mathrm{MSE}_g}$ over $G$
nonempty groups, not the mean of group RMSEs. Empty groups are reported
separately. MV is $b$ and SM is $n^{-1}\sum_{\mathcal V}d_id_i^{\mathsf T}$.
This exact decomposition describes statistical structure, not its physical
cause. Combining mean vectors across local frames requires a verified common
orientation. Otherwise, moments are reported separately for each frame, and
only rotation-invariant scalar quantities are aggregated.

\subsection{Predictive Uncertainty}
\label{sub:q3-trust}

An uncertainty region should balance coverage and size. This task evaluates
a symmetric positive-definite $2\times2$ Gaussian error matrix
$\widehat R_i$ around the reported position, rather than a scalar matching
score. It is the covariance parameter of the reported
zero-mean Gaussian error density, not necessarily the centered covariance of
the actual errors. Fixed-center likelihood fitting targets an error second
moment (Section~\ref{sec:compact-method}); it does not by itself imply nominal
chi-square coverage. We evaluate coverage empirically on reference residuals.
For the $n_R$ outputs with valid covariance estimates, define
\begin{equation}
 q_i=d_i^{\mathsf T}\widehat R_i^{-1}d_i,\quad
 \mathrm{NLL}_i=\tfrac12
 \left[q_i+\log\det\widehat R_i+2\log(2\pi)\right].
 \label{eq:nll}
\end{equation}
NLL evaluates the stated Gaussian density, including both scale and
orientation relative to the residual. With $c_\alpha=\chi^2_2(\alpha)$,
\begin{equation}
 \begin{aligned}
 \mathrm{PC}_\alpha&=n_R^{-1}\sum_i\mathbf1\{q_i\leq c_\alpha\},\\
 \mathrm{EA}_{\alpha,i}&=\pi c_\alpha\sqrt{\det\widehat R_i},\qquad
 \mathrm{DR}_{v,i}=\frac{(v^{\mathsf T}d_i)^2}
 {v^{\mathsf T}\widehat R_i v}.
 \end{aligned}
 \label{eq:predictive-diagnostics}
\end{equation}
Here $v$ is a predeclared unit direction. Coverage is checked at
$\alpha\in\{0.50,0.90,0.95\}$ against its nominal probability, not against
one. Ellipse areas quantify region size; directional ratios help identify
anisotropic mismatch. Nominal Gaussian interpretation requires a suitable
centered error model and reference. Unless otherwise stated, Q, NLL, EA and
DR are arithmetic means over outputs for which the metric is defined; their
sample counts accompany the availability scores.

\subsection{Controlled Position Update}
\label{sub:q4-update}

The same position can help or harm an update depending on the weight it is
given. The prior covariance $P$ is fixed before evaluation. An artificial prior error
$\xi$ has zero mean and covariance $P$, independently of the observation
error and its reported covariance. For
$K_i=P(P+\widehat R_i)^{-1}$, the updated error is
$(I-K_i)\xi+K_ie_i$, giving
\begin{equation}
 \begin{aligned}
 U_i(P)&=\operatorname{tr}[(I-K_i)P(I-K_i)^{\mathsf T}]
          +\|K_ie_i\|^2,\\
 \mathrm{UR}(P)&=N^{-1}\sum_iU_i(P),\qquad
 \mathrm{NR}(P)=\frac{\mathrm{UR}(P)}{\operatorname{tr}P}.
 \end{aligned}
 \label{eq:update}
\end{equation}
An explicit failure or rejection sets $K_i=0$, so rejecting every opportunity
gives NR$=1$. NR below one indicates lower expected squared error in this
controlled update, not a validated navigation benefit. Substituting $d_i$
for $e_i$ yields a reference-based proxy. The overall score requires a defined
risk for every evaluation case: unknown states, incompatible references, or
invalid covariances on successful outputs make it unavailable. Rejections
remain distinct from these unscorable cases. All comparisons use the same
prior and coordinate convention; success-only risk is reported as conditional.

For location $g$ with $N_g$ qualified opportunities, define
$\mathrm{UR}_g=N_g^{-1}\sum_{i\in g}U_i$ and
$\mathrm{NR}_g=\mathrm{UR}_g/\operatorname{tr}P$. The equal-location score is
\begin{equation}
 \mathrm{NR}_{\rm mac}(P)=\frac1{G_{\rm loc}}\sum_g\mathrm{NR}_g(P).
 \label{eq:macro-risk}
\end{equation}
Here $G_{\rm loc}$ counts locations; each retains failures with risk one.
This average weights each represented location equally, regardless of query
frequency. It describes the sampled locations rather than a broader geographic population.
Equation~\eqref{eq:update} instead weights by $N_g/N$. For an output-only score $s_i$ (NLL,
coverage indicator, or area), its macro summary is
$G_{\rm loc}^{-1}\sum_g n_g^{-1}\sum_{i\in\mathcal V_g}s_i$, where
$n_g=|\mathcal V_g|$. A location without outputs makes this summary unavailable.
The results and comparison criteria use these equal-location averages.
Unless otherwise stated, NR, NLL, PC and EA denote the macro quantities.

\subsection{Temporal Accumulation}
\label{sub:q5-temporal}

Repeated independent errors can average down; a persistent offset may not.
For observations with verified temporal order and equal spacing, we define
nonoverlapping within-flight windows $W$ of $m$ samples and duration $T=m\Delta t$:
\begin{equation}
 \mathrm{WME}(T)=\operatorname{mean}_{W}
 \left\|\frac1m\sum_{i\in W}e_i\right\|^2.
 \label{eq:temporal}
\end{equation}
WME distinguishes alternating from persistent errors even when their
marginal distance distributions coincide. $\mathrm{EL}_t$ records each
maximal above-threshold run length, and $\mathrm{ECnt}_t$ counts those runs.
Thresholds and windows are fixed in advance. Runs stop at gaps and flight
boundaries. Incomplete windows are counted but excluded from the metric,
without interpolation or concatenation. Reference residuals yield a proxy
for the corresponding true-error metric.
The empirical study uses unordered local localization tests, which support
neither temporal evaluation nor global map retrieval.

\section{Base-Anchored Uncertainty Calibration}
\label{sec:compact-method}
\subsection{Estimation Target}
The method estimates horizontal position uncertainty while keeping the
correspondences, DSM, position estimates and output availability unchanged.
Its effect is on observation weighting rather than localization accuracy.

Geometric sensitivity is computed from image and map perturbations, and
reference positions in the calibration set determine its scale without local
DSM-height ground truth. Uncertainty around the fixed reported position must
account for both dispersion and persistent bias. We therefore target the error
second moment rather than only the centered covariance. For a fixed reported
position, the Gaussian NLL optimum is the conditional second moment when that
moment is positive definite:
\begin{equation}
 R_*(x)=\mathbb E[ee^{\mathsf T}\mid x]
 =\operatorname{Cov}(e\mid x)+b(x)b(x)^{\mathsf T}.
 \label{eq:compact-moment}
\end{equation}
Here $x$ denotes available uncertainty features and $b(x)=\mathbb E[e\mid x]$.
This second moment includes persistent bias, whereas the centered covariance
describes dispersion around the mean error. They coincide when $b(x)=0$.
We approximate the second moment with a nonnegative model fitted by
equal-location-weighted likelihood; this restricted model need not attain the
unrestricted optimum.
Calibration and evaluation use the reference residual $d_i$
in place of unavailable true error $e_i$; coefficients are not physical DSM
variances. The configuration moment $A$ serves as a predictor feature for
the fitted error matrix.

In the model names, F denotes calibration on separate reference data;
G denotes geometry, D all-member
centered scatter, S selected-subset centered scatter, and A the base-anchored
moment. Iso denotes an isotropic variant; F-C is the constant full-matrix baseline.

\subsection{Label-Free Geometric Sensitivity}
For each successful pose, we stack the projection Jacobians of its RANSAC
inliers into $J$. Its columns correspond to camera-center increments in metres,
followed by three rotation increments in radians. With horizontal selector
$H=[I_2\;0]$, the first-order sensitivity is $L=HJ^\dagger$.
Jacobian columns are normalized for the rank test over all six pose
dimensions; physical units are then restored in the pseudoinverse.

Three independent, per-point unit perturbations define residual covariance
shapes $Q^{(s)}$: one query-pixel squared per image axis, one map-pixel squared
per map axis, and one metre squared in height. Map-pixel perturbations are
lifted onto the local surface, including horizontal--height cross terms. Their horizontal
position sensitivity matrix is
\begin{equation}
 G=\sum_{s\in\{\mathrm{query,map,height}\}}LQ^{(s)}L^{\mathsf T}.
 \label{eq:compact-propagation}
\end{equation}
The unit amplitudes fix the relative contribution of these three sensitivities;
one fitted coefficient does not identify their separate physical scales.
DSM finite differences describe a continuous local surface approximation,
not the exact derivative of the integer-pixel sampling operation.
This linearization excludes selection uncertainty, cross-point map correlation
and distant alternative poses. When surface slopes are nonfinite or the rank
test fails, the model uses the F-C fallback defined below.

\subsection{Configuration Disagreement}
The base localizer uses SIFT with at most 8192 features, a 0.80 ratio test, and
DSM-backed SQPnP RANSAC (4-pixel reprojection threshold, 200 iterations, confidence 0.999).
At least 12 finite correspondences and eight inliers are required; rejected
or failed solves are included as cases without a position estimate.
Four further configurations independently change the ratio to 0.75 or 0.85,
halve the feature budget, or inset the map by 10\%. All use the same inputs.
For $m_i\geq3$ successful configurations, including the base, their centered scatter is
\begin{equation}
 D_i=m_i^{-1}\sum_j(z_{ij}-\bar z_i)(z_{ij}-\bar z_i)^{\mathsf T}.
 \label{eq:compact-disagreement}
\end{equation}
This is sensitivity to selected configurations, not an ensemble of independent
measurements, since all configurations may share an error. Computing this
feature requires four additional localizer runs, unlike the constant baselines.

\subsection{A Moment Anchored to the Reported Position}
Configuration spread need not reflect the error of the reported position.
A single inconsistent variant can make $D_i$
large while the base is accurate. Conversely, removing that variant without
accounting for the base can obscure its displacement from the consistent
group.

We therefore retain the base $z_{i0}$ and use a base-relative moment feature.
For $m_i\geq3$ successful configurations, we set $k_i=\lfloor m_i/2\rfloor+1$ and select
the $k_i$-member subset $S_i$ with minimum maximum pairwise Euclidean distance.
Exact ties are resolved lexicographically by configuration name. This selection objective is
the same as in Minimum-Diameter Averaging~\cite{elmhamdi2020mda}. We use the
selected members to form a moment around the unchanged base:
\begin{align}
 A_i&=k_i^{-1}\sum_{j\in S_i}(z_{ij}-z_{i0})(z_{ij}-z_{i0})^{\mathsf T}
 \nonumber\\
 &=C_{S_i}+(\bar z_{S_i}-z_{i0})(\bar z_{S_i}-z_{i0})^{\mathsf T},
 \label{eq:anchored-moment}
\end{align}
where $C_{S_i}$ is the subset's centered scatter with divisor $k_i$.
The second term preserves disagreement with an isolated base. The resulting
$A_i$ describes configuration displacements; the subsequent calibration maps
this feature to a predicted base error moment within a restricted model family.

The base-anchored model, F-GA, replaces $D_i$ by $A_i$ in
Eq.~\eqref{eq:compact-gdc}, retaining the same $G_i$, base, inputs and solver
budget. F-A omits geometry. F-GS uses $C_{S_i}$, and F-GA-all instead uses
the all-member moment around the base; each is fitted independently on the
same calibration samples. These comparisons separate subset selection from
anchoring. F-GA-Iso preserves the trace of F-GA without refitting.

A majority of successful configurations may still be inaccurate. Such a majority can
downweight an accurate base; a common displacement of all members leaves every
relative feature unchanged. Zero scatter can therefore coexist with nonzero
error. Although the common term and geometry supply additional information,
these failure modes remain possible. When fewer than three configurations
succeed, the method uses the F-C fallback. Equivariance tests cover similarity-coordinate
transformations; general affine equivariance remains unverified.

\begin{figure*}[t]
\centering
\includegraphics[width=7.1in]{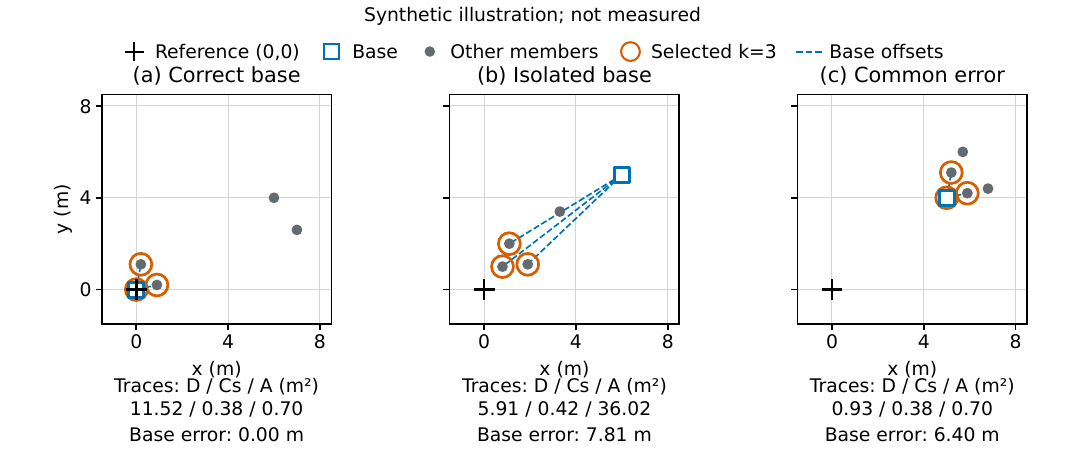}
\caption{Synthetic operator examples on identical metre-scale axes.
Cross: reference $(0,0)$; square: fixed base; rings: minimum-diameter subset
($k=3$); dashed segments: base-relative offsets. Traces of all-member scatter
$D$, subset scatter $C_S$ (Cs), and anchored moment $A$ are computed from
the displayed configurations. (a) Deviating variants inflate $D$. (b) $A$ retains an isolated
base's offset from a compact subset. (c) Shared displacement remains invisible.
Matrices describe raw configuration dispersion rather than calibrated error
covariances. The reference is shown for interpretation and is excluded from
subset selection and scatter calculations.}
\label{fig:base-anchored-diagram}
\end{figure*}

Figure~\ref{fig:base-anchored-diagram} illustrates the three configuration
statistics on synthetic positions with common metre-scale axes: inconsistent
variants, an isolated base and a displacement shared by all configurations.
The matrices describe raw configuration dispersion rather than calibrated
error covariances.

\subsection{Calibration Models and Baselines}
Let $w_i$ give every nonempty fitting location equal total weight, with
$\sum_iw_i=1$. The floor-regularized isotropic (F-I) and full-matrix (F-C) controls are
\begin{equation}
 R_{\rm I}=\epsilon I_2+\tfrac12\sum_iw_i\|d_i\|^2I_2,\quad
 R_{\rm C}=\epsilon I_2+\sum_iw_i d_i d_i^{\mathsf T}.
 \label{eq:compact-common}
\end{equation}
For a configuration feature $V_i$, the combined model predicts
\begin{equation}
 R_i=\epsilon I_2+u(\alpha I_2+\beta\widetilde G_i+
 \gamma\widetilde V_i),\qquad \alpha,\beta,\gamma\geq0,
 \label{eq:compact-gdc}
\end{equation}
where $\epsilon=10^{-6}$ m$^2$ is the regularization floor and $u$ is half
the mean squared reference error in the calibration set, bounded below by
that floor. Each feature matrix is normalized by half its mean trace on
the same set; these normalization factors remain fixed during evaluation.
Coefficients minimize group-weighted Gaussian NLL using deterministic
multi-start L-BFGS-B, at most 500 iterations per start.
F-GA uses $V=A$; F-GD uses $V=D$. F-G, F-D and F-A each have
two coefficients and three starts; joint models have three coefficients and
four starts. This is a fixed per-start budget, not equal total fitting cost.
F-C also has three fitted matrix entries.
Section~\ref{sec:frozen-parameters} gives the fitted parameters and convergence
results. All parameters are estimated on the calibration set.

F-I and F-C use all 262 successful calibration samples; Config and Config-T use
the 260 with valid disagreement. All component and joint models use exactly
the same 248 calibration samples with complete features. Iso controls replace a valid joint matrix by half its trace
times $I_2$, without refitting; this tests overall directional shape, not
off-diagonal correlation alone. Missing required shapes invoke the same fixed
F-C fallback, including for the isotropic variants. The results include fallback counts.
Because F-C and the feature models use different calibration samples, their
contrast compares complete calibration procedures, not an isolated feature
ablation. Component comparisons on the same 248 samples and the H-Q control
address the narrower question of feature contribution.

Two additional baselines calibrate configuration scatter directly. Config fits
$a+b\sqrt{\operatorname{tr}D_i}$ to the reference-residual radius by weighted least squares
and squares the predicted radius, floored at $10^{-3}$ m, to obtain isotropic
variance. Config-T multiplies that matrix by the Gaussian-likelihood
temperature $t=\sum_iw_i d_i^{\mathsf T}R_i^{-1}d_i/2$, estimated on the calibration set.
This temperature rescales the covariance while preserving the position and
matrix shape.
The formulation draws on uncertainty-aware geometric
estimation~\cite{vakhitov2021uncertainty}, multivariate covariance
recalibration~\cite{kuppers2022parametric}, and fixed-mean scalar
recalibration~\cite{levi2020evaluating}. The nonnegative additive feature
combination in Eq.~\eqref{eq:compact-gdc} is the restricted model family used
in this study. Its distinctive element is the
base-relative feature and its combination with matching/map sensitivity,
evaluated through both predictive quality and controlled position updates.

\subsection{Fitted Parameters}
\label{sec:frozen-parameters}
Table~\ref{tab:frozen-parameters} reports the parameters used in evaluation.
All seven feature-based models use the same 248 samples with complete features at
24 equally weighted locations. In Eq.~\eqref{eq:compact-gdc}, the common
constants are $\epsilon=10^{-6}\,\mathrm{m}^2$,
$u=1.14822827\,\mathrm{m}^2$ and $s_G=2.06950176\,\mathrm{m}^2$.
Other mean-half-trace divisors are $s_D=4.12298340$, $s_A=0.893085303$,
$s_S=0.314136125$ and $s_{A_{\rm all}}=4.59884867$, all in
$\mathrm{m}^2$; $S$ denotes the selected-subset centered scatter.
Coefficients are dimensionless.

\begin{table*}[!t]
\centering
\caption{Fitted model coefficients and convergence. All 25 starts converged
within the 500-iteration limit; the solution with the lowest calibration
objective was retained. Values are rounded for display.}
\label{tab:frozen-parameters}
\begin{tabular}{@{}llrrrrr@{}}
\toprule
Model & $V$ & $\alpha$ & $\beta$ & $\gamma$ & Converged/starts & Iteration range \\
\midrule
F-G & --- & 0 & 2.252799 & --- & 3/3 & 7--11 \\
F-D & $D$ & 0.018637 & --- & 11.790664 & 3/3 & 15--20 \\
F-A & $A$ & 0.025794 & --- & 3.250298 & 3/3 & 14--17 \\
F-GD & $D$ & 0.008367 & 0.289716 & 5.900141 & 4/4 & 22--33 \\
F-GA & $A$ & 0.013870 & 0.203725 & 2.403562 & 4/4 & 21--26 \\
F-GS & $S$ & 0.006267 & 0.442200 & 2.484152 & 4/4 & 19--26 \\
F-GA-all & $A_{\rm all}$ & 0.008668 & 0.237032 & 4.566602 & 4/4 & 22--25 \\
\bottomrule
\end{tabular}
\end{table*}

The two constant baselines use all 262 successful calibration samples, giving
\begin{equation}
 \begin{aligned}
 R_{\rm I}&=1.150871937I_2,\\
 R_{\rm C}&=\begin{bmatrix}
 1.592216550&-0.282244940\\-0.282244940&0.709527324
 \end{bmatrix}\quad(\mathrm{m}^2).
 \end{aligned}
\end{equation}
These matrices already include the floor. Config uses 260 disagreement-valid
records and the radius map $0.232038420\,\mathrm m+
0.449170138\sqrt{\operatorname{tr}D}$; Config-T multiplies its squared-radius
matrix by $0.651679894$. All fallbacks use the same $R_{\rm C}$, including
the isotropic variants. Numerical convergence alone guarantees neither a
global optimum nor generalization.

\section{Experimental Protocol}
\label{sec:compact-results}
\begin{table*}[!t]
\centering\small
\caption{Experiment stages and evaluation data. Counts denote
opportunities unless explicitly marked as outputs. Later exploratory studies
reuse the earlier data.}
\label{tab:evidence-stages}
\begin{tabular}{@{}p{0.20\textwidth}p{0.24\textwidth}p{0.22\textwidth}p{0.27\textwidth}@{}}
\toprule
Study & Calibration data & Evaluation data & Interpretation \\
\midrule
Ranking diagnostic & 246 outputs, 3 locations & 247 outputs, same 3 locations & Within-location development analysis; overall update risk unavailable \\
Method development & 300 opportunities, 24 locations & 258 opportunities, disjoint locations & F-GD failure analysis motivating anchoring \\
Main comparison & Original 300; 248 complete-feature outputs & 728 validation and 142 inPlace; 24 and 3 represented locations & Held-out queries; thirteen models fixed before reference labels were accessed \\
Base-quality control & Original 248 complete-feature outputs & Same 728/142, previously evaluated & Exploratory H-Q comparison with no extra localizer runs \\
Conditional addition & Nested location holdout over original 300; calibration-only fitting & 300 source OOF plus previously evaluated 728/142 & Exploratory H-QA/H-QI comparison within the same domain \\
\bottomrule
\end{tabular}
\end{table*}
Table~\ref{tab:evidence-stages} distinguishes method development, the main
evaluation and subsequent exploratory analyses.
OOF denotes out-of-fold predictions. The base-anchored model and its primary
comparisons were specified before the evaluation labels were accessed.
Subsequent analyses reuse these data and are therefore exploratory.
\begin{figure*}[t]
\centering
\includegraphics[width=\textwidth]{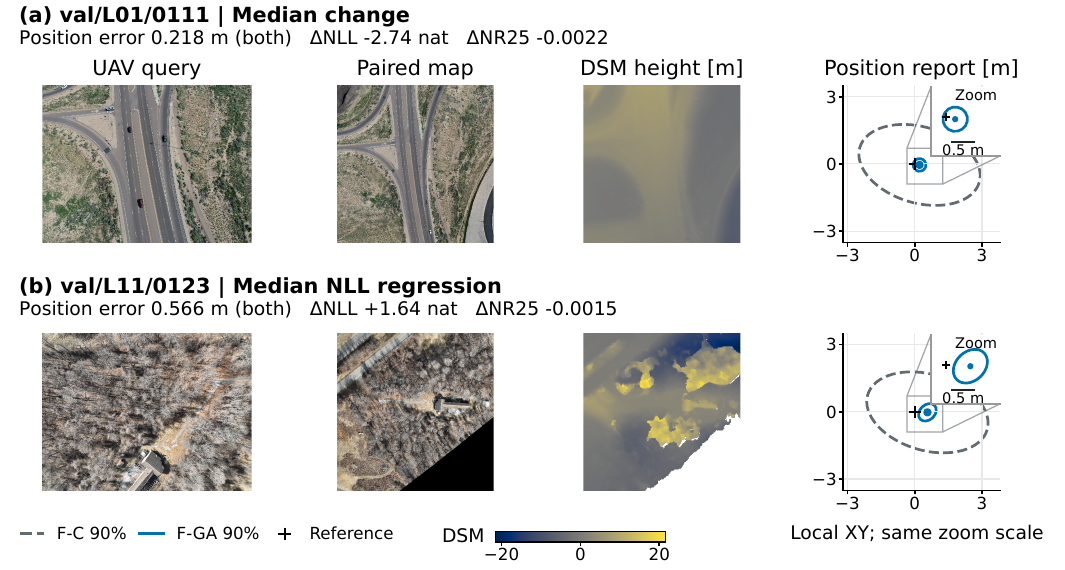}
\caption{Real inputs and the uncertainty attached to an unchanged position.
Rows show the cases with median F-GA-minus-F-C NLL change among all 629
validation outputs and among the 33 cases with increased NLL, respectively;
ties are resolved by query ID.
Query images, paired maps and DSM heights are from OrthoLoC.
Heights share a color scale and represent the surface, not DSM error;
blank map areas are retained. The right column shows dataset-local XY
coordinates relative to the reference position. The dot is the shared prediction;
both 90\% regions share this center.
Linked insets share limits and a 0.5~m scale bar. The tighter F-GA regions contain the reference in (a)
but miss it in (b), despite lower NR25 in both. The examples illustrate
individual outcomes alongside the complete-cohort results.}
\label{fig:anchored-real-cases}
\end{figure*}

\subsection{Data and References}
The experiment uses the OrthoLoC release's R conditions,
whose maps derive from acquisition-image reconstruction
\cite{dhaouadi2025ortholoc}. Inference reads only the query image, paired map
image, DSM XYZ raster, metric pixel scale and intrinsics. It excludes reference
extrinsics, pose-derived point maps and meshes. The released
world-to-camera transform defines a reference camera center $-R^{\mathsf T}t$;
its horizontal coordinates are dataset-local metric XY, not geodetic ENU.
All experiments use the raster's stored metric XYZ coordinates and per-file XY offset.
Unknown reference uncertainty limits physical accuracy claims.

Figure~\ref{fig:anchored-real-cases} shows actual inputs and reported regions:
their centers stay fixed, while uncertainty changes.

Calibration uses 300 development queries from 24 locations in the official
training split, including reruns of queries that had previously failed.
An initial evaluation uses all 258 R conditions at validation locations
L27--L49, which are absent from the calibration set. Its outcome motivates the anchored
feature, so these 258 conditions are development evidence for that feature,
not its confirmation set.

The main evaluation uses the same calibration set and two
predefined query sets:
all remaining official validation R conditions, and all official test\_inPlace
R conditions. Required inputs are available for 728 of 772 validation conditions
at 24 locations. The remaining 44 L06 records lack metric scale and are excluded
before localization rather than counted as localization failures. All 142
inPlace conditions at L01, L02 and L15 have the required inputs. Every included
location is represented in the calibration set. Validation and inPlace results are
reported separately; they are query-held-out, not unseen-location tests.
Exact hashes of query arrays across all 8614 R records identified no duplicate
queries within the evaluation sets or between these sets and previously
used queries. Near-duplicate views and shared maps remain possible.
Predictions from all 13 models were fixed for both sets before their reference
labels were accessed.

All experiments use paired local map crops from one map-production domain.
The related xDOP/xDOPDSM conditions use external maps\cite{dhaouadi2025ortholoc}
and are excluded from both evaluation and auxiliary inputs. The shared
reconstruction of query imagery and maps in R may simplify localization.
Robustness to arbitrary DSM bias remains untested.

\subsection{Comparison Criteria}
All covariance models use identical positions, inliers, success/failure states and
random seeds. The code and runtime environment are fixed.
Each worker uses one OpenCV thread with OpenCL disabled, and each query uses
the same random seed in serial and parallel execution. Coefficients,
fallbacks and models use only calibration data.

The prespecified criteria combine an equal-location NLL improvement of at least
0.05 nat over F-C with no increase in overall NR at $P=25I_2$ m$^2$,
and lower NLL than both single-component models on queries with complete
features for all compared models.
For F-GA, two additional comparisons require NR25 no greater than F-GD or
F-GA-all on the validation set. F-GS isolates anchoring; the Iso control tests
directional information separately. Validation is the primary evaluation;
the same six comparisons are reported separately for inPlace as a
corroborating evaluation. F-GA was specified before evaluation,
and the baselines were reused without refitting. Secondary
priors and ablations assess how the results vary beyond the primary comparisons.

We report each location and equal-location aggregates, retaining explicit
failures with prior-only update risk. The covariance models share identical
position statistics. PC is checked at 50\%, 90\% and 95\%; update priors
are $4I_2$, $25I_2$ and $100I_2$ m$^2$. We use NR25 as shorthand for
$\mathrm{NR}_{\rm mac}(25I_2\,\mathrm{m}^2)$, and analogously NR4/NR100.
A paired bootstrap resamples each
evaluation set's locations 10,000 times with seed 20260930. These intervals
describe variation across the sampled locations, not geographic generalization;
this distinction is particularly important for the three-location inPlace set.
Data splits, metrics and coefficients were held fixed for each evaluation.

\subsection{Numerical Verification}
Figure~\ref{fig:compact-counterexamples} isolates group offsets, covariance
and temporal ordering. We test the exact moment identity and verify the
independent-prior update with 200,000 fixed-seed samples and 2\% tolerance.
The tests include noncommuting matrices and all-rejection risk, and check the
handling of invalid covariances, missing references, duplicate identities and
unknown states.

\section{Results}
\label{sec:results}
\begin{figure*}[t]
\centering
\includegraphics[width=\textwidth]{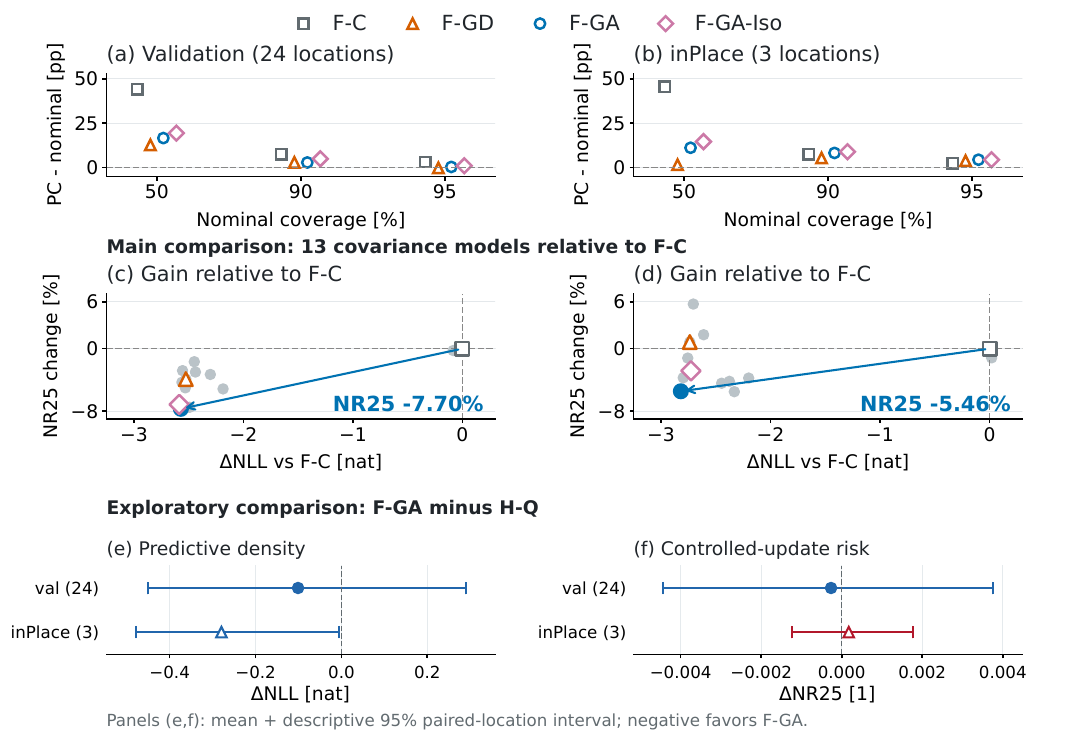}
\caption{Calibration, controlled updates and comparison with base-only quality.
(a,b) Coverage minus nominal level; zero is ideal. Markers are offset
horizontally for legibility.
(c,d) Thirteen covariance models relative to F-C: NLL difference and percentage
NR25 change. Arrows connect F-C to F-GA; negative values on both axes indicate
improvement in both metrics.
(e,f) Exploratory comparisons with H-Q on the same evaluation sets:
F-GA minus H-Q, with descriptive 95\%
paired-location bootstrap intervals (10,000 resamples, seed 20261009).
Negative favors F-GA; both validation intervals cross zero and inPlace NR25
slightly favors H-Q. All panels use equal-location means: 24 validation and
three inPlace locations. PC/NLL include 629/136 outputs with
fallbacks; NR25 includes all 728/142 eligible queries with no-output risk one,
under $P=25I_2$ m$^2$. Positions and states are unchanged. The intervals
describe these locations rather than a broader geographic population.}
\label{fig:anchored-quality-use}
\end{figure*}

\subsection{Availability and Position Error}
All 870 conditions with the required inputs were evaluated. Validation produces 629
positions and 99 failures; inPlace produces 136 positions and six failures.
All 13 models use these same positions and failure states.
Among successful outputs, the median distances are 0.189 and 0.163 m, with micro RMSE 1.410
and 0.445 m, respectively. The maximum validation error is 24.771 m.

\begin{table}[t]
\centering\fontsize{9}{11}\selectfont
\setlength{\tabcolsep}{3pt}
\caption{Equal-location averages of position-error moments (m$^2$) among successful outputs. $b_g$ is the location's mean residual and $W_g=\operatorname{mean}_{i\in g}\|d_i-b_g\|^2$. Mean $\|b_g\|^2$ summarizes local offsets, not global OSM. Global BGM also requires a verified common orientation across locations. All thirteen models use the same positions.}
\label{tab:anchored-groups}
\begin{tabular}{lrrr}
\toprule
Cohort & MSE$_{\rm mac}$ & Mean $\|b_g\|^2$ & Mean $W_g$ \\
\midrule
Val & 1.5649 & 0.0430 & 1.5218 \\
InPlace & 0.3809 & 0.0284 & 0.3525 \\
\bottomrule
\end{tabular}
\end{table}

Within-location dispersion dominates the squared location mean
(Table~\ref{tab:anchored-groups}). The data support evaluation of availability,
position agreement, within-location error structure, predictive uncertainty
and controlled updates. Global OSM/BGM require a verified common orientation
across locations, and temporal accumulation requires ordered observations;
neither is available here.

Validation failures comprise 87 cases with no PnP consensus, 11 with fewer
than eight inliers, and one explicit solver exception. InPlace has three
no-consensus and three insufficient-inlier failures. Each failure contributes
prior-only risk to the overall score. F-GA has 585 validation outputs with
complete features and 44 F-C fallbacks: 31 cases with nonfinite DSM slopes and
13 insufficient-member cases. InPlace has 132 complete outputs and four
fallbacks: three cases with nonfinite DSM slopes and one with insufficient members.

\begin{table*}[t]
\centering\fontsize{9}{11}\selectfont
\setlength{\tabcolsep}{3.5pt}
\caption{Comparison of thirteen covariance models using equal-location means, identical position estimates and identical failure states. NLL, PC90 and EA90 use successful outputs; NR25 includes every eligible query, with risk one for missing outputs. Bold cells mark minimum NLL/NR25 or PC90 closest to 0.90, based on unrounded values. EA90 describes region size and is interpreted alongside coverage. $\dagger$ denotes an NR25 tie after rounding only.}
\label{tab:anchored-summary}
\begin{tabular}{lrrrrrrrr}
\toprule
& \multicolumn{4}{c}{Validation} & \multicolumn{4}{c}{inPlace} \\
\cmidrule(lr){2-5}\cmidrule(lr){6-9}
Method & NLL $\downarrow$ & PC90 & EA90 & NR25 $\downarrow$ & NLL $\downarrow$ & PC90 & EA90 & NR25 $\downarrow$ \\
& (nat) & target 0.90 & (m$^2$) & & (nat) & target 0.90 & (m$^2$) & \\
\midrule
\multicolumn{9}{l}{\textit{Constant and configuration-only baselines}} \\
F-I & 2.6583 & 0.9763 & 16.6503 & 0.1549 & 2.1439 & 0.9744 & 16.6503 & 0.0410 \\
F-C & 2.7420 & 0.9739 & 14.8253 & 0.1552 & 2.1251 & 0.9744 & 14.8253 & 0.0415 \\
Config & 0.4404 & 0.9549 & 29.8456 & 0.1501 & -0.0748 & 0.9744 & 3.1150 & 0.0399 \\
Config-T & 0.3041 & \textbf{0.9131} & 19.5825 & 0.1506 & -0.2508 & 0.9548 & 2.0607 & 0.0398 \\
\addlinespace[2pt]
\multicolumn{9}{l}{\textit{Geometry and centered-disagreement controls}} \\
F-G & 0.5566 & 0.9602 & 7.7931 & 0.1473 & -0.2057 & 0.9744 & 2.7347 & \textbf{0.0392}$^{\dagger}$ \\
F-D & 0.1866 & 0.9279 & 13.5486 & 0.1509 & -0.5802 & 0.9513 & 1.4922 & 0.0439 \\
F-GD & 0.2167 & 0.9292 & 9.6063 & 0.1492 & -0.6125 & 0.9537 & 2.0613 & 0.0418 \\
F-GD-Iso & 0.2938 & 0.9385 & 112.1950 & 0.1527 & -0.4861 & \textbf{0.9341} & 7.1488 & 0.0422 \\
\addlinespace[2pt]
\multicolumn{9}{l}{\textit{Base-anchored model and component comparisons}} \\
F-A & 0.2504 & 0.9177 & 6.0468 & 0.1436 & -0.6723 & 0.9685 & 1.5978 & 0.0399 \\
\textbf{F-GA} (proposed) & 0.1712 & 0.9284 & 7.1256 & \textbf{0.1433} & \textbf{-0.6941} & 0.9821 & 1.8579 & 0.0392$^{\dagger}$ \\
F-GS & 0.2119 & 0.9352 & 6.4426 & 0.1475 & -0.3216 & 0.9453 & 1.1919 & 0.0397 \\
F-GA-all & 0.1807 & 0.9311 & 10.5228 & 0.1486 & -0.6311 & 0.9537 & 2.0427 & 0.0410 \\
F-GA-Iso & \textbf{0.1557} & 0.9491 & 28.7510 & 0.1442 & -0.6021 & 0.9888 & 8.3313 & 0.0403 \\
\bottomrule
\end{tabular}
\end{table*}

\subsection{Predictive Density and Region Calibration}
F-GA lowers validation NLL relative to F-C. F-C uses 262 calibration samples, whereas F-GA
uses 248; this comparison therefore measures the complete calibration
procedures rather than the feature effect alone.
On validation, the F-GA minus F-C NLL difference is $-2.5708$ nat, with
descriptive paired-location interval $[-3.3259,-2.0126]$; 23 locations improve
and L04 regresses. Its 90\% regions are smaller and coverage is closer to
nominal (Table~\ref{tab:anchored-summary}). At nominal 50/95\%, F-GA covers
66.60/95.33\%, so calibration remains imperfect.
Figure~\ref{fig:anchored-quality-use} compares coverage at three probability
levels, NLL and update risk.

On inPlace, F-GA also lowers NLL and reduces region area, but 90\% coverage changes
from 97.44\% to 98.21\%, farther from nominal. PC measures coverage of reference residuals rather
than errors relative to the unknown true camera center.

\subsection{Component and Mechanism Comparisons}
On validation queries with complete features for all compared models,
F-GA has NLL $-0.1329$, versus
0.0895 for F-A and 0.4195 for F-G, meeting both component criteria.
On all outputs, including F-C fallbacks, gains are more modest against F-A: the paired NLL
difference is $-0.0792$ nat, with interval $[-0.2547,0.0540]$. The result
suggests that geometry helps on the complete-feature subset, while its
contribution is less certain when fallback cases are included.

\begin{figure*}[t]
\centering
\includegraphics[width=\textwidth]{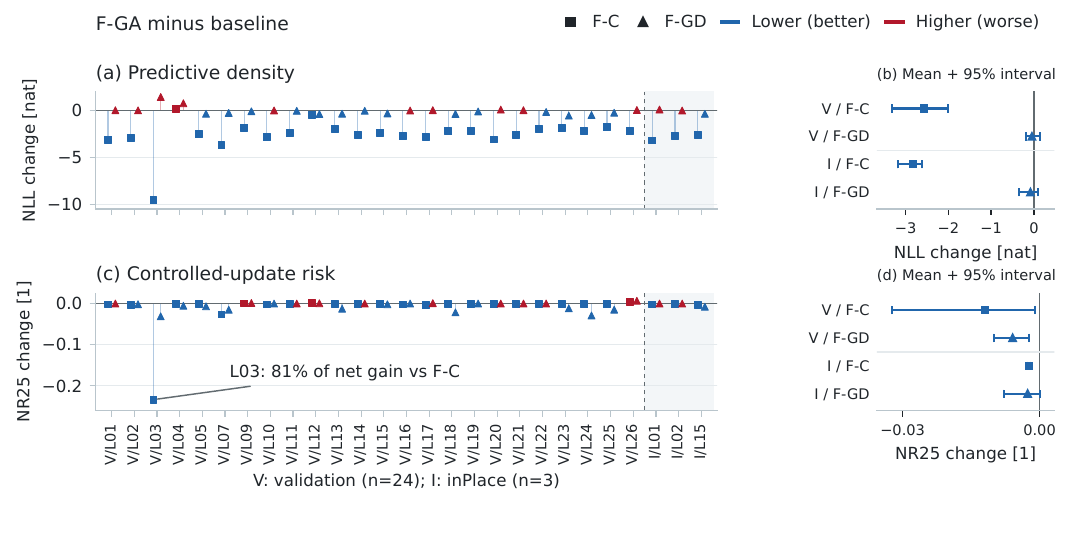}
\caption{Paired location differences, F-GA minus F-C (squares) or F-GD
(triangles). Blue negative values indicate improvement; red positive values
indicate regression. The 27 cohort--location entries share
a fixed order; dashed lines separate validation (V) and inPlace (I).
Right: means and descriptive 95\% paired-location bootstrap intervals for
each cohort and comparator. Resampling is by location rather than query;
inPlace has only three locations. The right panels use different x-axis
limits from the left panels; scales are shared within each panel.
NLL uses outputs with valid covariance estimates; NR25 uses all eligible queries
(risk 1 for cases without an output). Position predictions are unchanged.}
\label{fig:anchored-site-deltas}
\end{figure*}

The corresponding centered-subset F-GS and all-member anchored F-GA-all
scores are $-0.0878$ and $-0.1256$ on the same queries. Both controls are
fitted independently with the same budget and have higher NLL point estimates. The all-output
NLL differences against F-GS, F-GA-all and F-GD have descriptive intervals
that include zero. Figure~\ref{fig:anchored-site-deltas} shows F-C and F-GD
contrasts by location, alongside descriptive intervals for each evaluation set.

F-GA-Iso has lower validation NLL than F-GA (0.1557 versus 0.1712), although
its mean 90\% area is much larger (28.751 m$^2$). The anisotropic model
therefore offers a more compact region, but not a directional NLL advantage
on the validation set. The opposite NLL ordering on inPlace indicates that
this trade-off depends on the evaluation conditions.

\subsection{Controlled Update Performance}

F-GA reduces NR25 relative to F-C by 7.70\% on validation and 5.46\% on
inPlace, including failures. Both evaluation sets meet all six prespecified
comparisons with the constant baseline, individual components and alternative
configuration models; the absolute scores are shown in
Table~\ref{tab:anchored-summary}. These controlled updates assume an independent prior;
real prior--observation correlation would introduce an additional cross term.

On validation, paired-location NR25 intervals are $[-0.03240,-0.00097]$
against F-C and $[-0.01004,-0.00226]$ against F-GD. F-GA improves
21 of 24 locations against F-C, but L03 supplies about 81\% of the net
aggregate gain. Against F-GS, the interval $[-0.01164,0.00013]$ includes
zero, leaving the incremental benefit of anchoring uncertain. InPlace intervals against F-GD,
F-GS and F-GA-all also include zero.

\begin{figure*}[t]
\centering
\includegraphics[width=\textwidth]{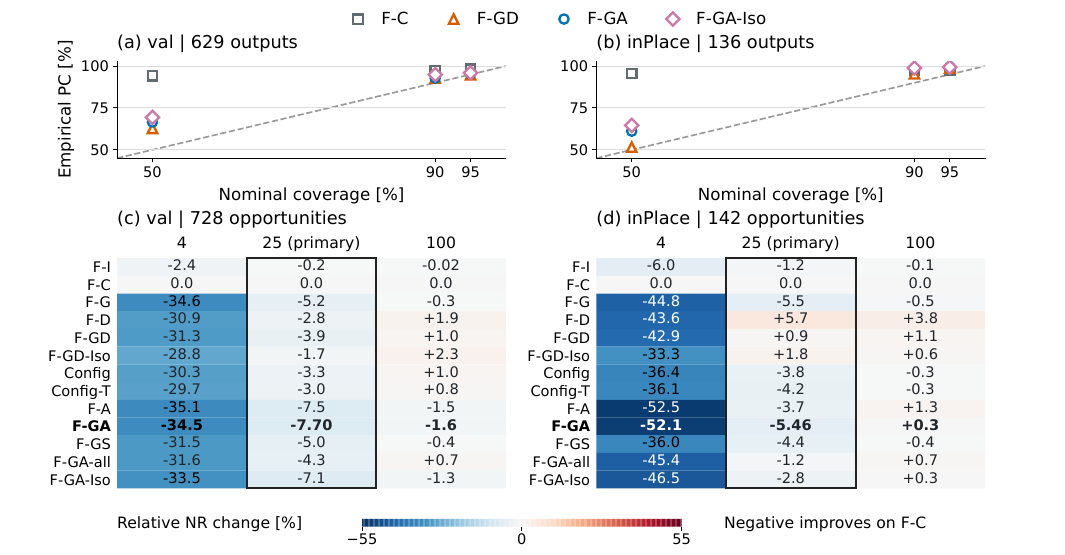}
\caption{Coverage and update risk across prior settings. (a,b) Coverage at nominal
50\%, 90\%, and 95\%; diagonals denote ideal coverage. (c,d) All thirteen
models under $P=pI_2$, $p\in\{4,25,100\}\,\mathrm{m}^2$ (outlined:
primary $p=25$). Cells show
$100[\mathrm{NR}_{m}(p)/\mathrm{NR}_{\mathrm{F-C}}(p)-1]\%$; negative
values indicate lower risk. Heatmaps share a symmetric scale. Equal-location
means use 24 validation and three inPlace locations. Coverage uses 629/136
successful outputs; NR includes all 728/142 eligible cases, with no-output risk
one. Performance ordering varies with the prior.}
\label{fig:cross-prior-complete}
\end{figure*}

The model ordering changes with the prior and location.
InPlace F-G is slightly better at NR25
(0.03920), and F-GA's NR100 (0.03437) is slightly worse than F-C's
(0.03426). On validation at $P=4I_2$, F-A and F-G are also below F-GA.
Figure~\ref{fig:cross-prior-complete} compares all three prior settings
using the same fitted models.
At query level, validation L26/0068 has NR25 0.23527 under F-GA
versus 0.12091 under F-GD; for L03/0056, NLL rises from 3.10 under F-GD to
43.46 under F-GA.

Among successful queries, the median runtimes for the base solve and the
four additional variants are
1.867 and 6.351 s on validation, and 2.162 and 7.228 s on inPlace, respectively.
Measurements used a shared CPU environment with eight concurrent
single-threaded workers for validation and two for inPlace. These offline
timings do not establish real-time performance; configuration-based models
incur computation that the constant baselines avoid.

\section{Exploratory Control Studies}
\label{sec:exploratory-controls}
We compare F-GA with H-Q, a calibration model that uses base-solve cues without
additional localizer runs. We also test the incremental value of the anchored
feature within the H-Q model family through a conditional addition. Both studies
assess whether extra configurations provide information beyond the base solve,
with every position and failure state unchanged. We designed these exploratory
studies after inspecting the earlier results.

\subsection{Base-Only Quality Calibration}
\label{sec:quality-control}
H-Q uses the same 248 calibration samples with complete features at
24 locations, without running additional localizer configurations.
Given the fitted F-G error matrix $R_G$, it predicts
\begin{equation}
 R_H=\exp[\operatorname{clip}(w^{\mathsf T}\phi,-20,20)]R_G+10^{-6}I_2.
 \label{eq:quality-report}
\end{equation}
The four raw cues are $\log n_{\rm inlier}$,
$\log\max(\mathrm{RMS}_{\rm reproj},10^{-6})$,
$\log\max(\sqrt{\det C_q},10^{-12})$, and
$\log(s_{\max}/s_{\min})$. Here $C_q$ is the population covariance of
inlier coordinates transformed by $K_{\rm cam}^{-1}$, where
$K_{\rm cam}$ is the camera-intrinsic matrix, and $s$ are the singular
values of the column-normalized pose Jacobian used for geometric sensitivity.
The cues describe the number of correspondences, reprojection error, spatial
extent and geometric conditioning. They predict error scale empirically rather
than specify a physical noise model.

The location-weighted cue means in the calibration set are
$(5.125256,-0.059649,-2.766093,3.081946)$ and standard deviations are
$(0.865389,0.160479,0.391677,0.366263)$.
The feature vector $\phi$ contains an intercept, four standardized cues clipped to
$[-5,5]$, and four missing-value indicators. Missing cues are replaced by
their calibration means rather than excluded. All calibration cues are
observed, so the four missing-value coefficients are zero. The intercept
and cue coefficients are
$(-0.582925,-0.108810,0.471415,-0.002832,0.085328)$.
Performance when cues are missing has not been validated.

The objective is equal-location Gaussian NLL plus
$\lambda\|w_{\rm nonintercept}\|^2/2$. Leave-one-location-out validation
within the calibration set selects $\lambda=1$ from $\{0,0.01,0.1,1,10\}$; respective
mean validation NLL values are 0.704390, 0.700115, 0.673471, 0.632617 and
0.934455. Each fold refits F-G and cue statistics on its remaining locations.
The final H-Q model uses F-G fitted on the full calibration set.
When geometry is unavailable, it uses F-C: 31 of 629 validation outputs and
3 of 136 inPlace outputs. The localization failures remain 99 and 6.

\begin{table*}[t]
\centering\small
\caption{Exploratory comparison with base-only calibration on the same evaluation sets.
Equal-location PC90 and EA90 use all outputs including fallbacks; NR25 retains
all eligible queries. Results are reported separately for validation and
inPlace; the latter contains only three locations.}
\label{tab:quality-control}
\begin{tabular}{@{}llrrrr@{}}
\toprule
Cohort & Model & NLL (nat) & PC90 (\%) & EA90 (m$^2$) & NR25 \\
\midrule
Val & F-C & 2.7420 & 97.39 & 14.825 & 0.155248 \\
Val & H-Q & 0.2722 & 93.14 & 6.996 & 0.143556 \\
Val & F-GA & 0.1712 & 92.84 & 7.126 & 0.143292 \\
InPlace & F-C & 2.1251 & 97.44 & 14.825 & 0.041488 \\
InPlace & H-Q & -0.4146 & 88.76 & 2.776 & 0.039049 \\
InPlace & F-GA & -0.6941 & 98.21 & 1.858 & 0.039224 \\
\bottomrule
\end{tabular}
\end{table*}

On validation, F-GA minus H-Q NR25 is
$-0.000264$, with descriptive paired-location interval
$[-0.004429,0.003751]$ (Table~\ref{tab:quality-control},
Fig.~\ref{fig:anchored-quality-use}(e,f)). Excluding L03 reverses its sign to $+0.001230$;
the main result includes that location. On inPlace, F-GA's NR25
is slightly higher, with difference $+0.000175$ and interval
$[-0.001244,0.001769]$. H-Q's PC90 is 93.14/88.76\%, versus F-GA's
92.84/98.21\%. The intervals leave the incremental
update benefit over this less expensive model uncertain; they do not
establish equivalence.

Intervals use 10,000 paired-location bootstrap resamples with seed 20261009.
They describe the sampled locations and do not account for variation across
independent map-production domains. This exploratory comparison reuses the
earlier evaluation data.

\subsection{Adding the Anchored Feature to Base-Only Calibration}
\label{sec:conditional-anchor}
The conditional-addition study tests whether the anchored feature adds
information to H-Q through an additive error term. Its design was specified
before the new fits and scores, but the calibration and evaluation observations
had already been evaluated.
Given the H-Q error matrix $R_H$, we compare two additive models:
\begin{equation}
 \begin{aligned}
 R_{HA}&=R_H+c_AuA/s_A,\qquad c_A\geq0,\\
 R_{HI}&=R_H+c_IuI_2,\qquad c_I\geq0.
 \end{aligned}
\end{equation}
H-QA adds the anchored feature, whereas H-QI adds a constant isotropic term.
Each model introduces one coefficient,
fitted by equal-location Gaussian NLL from initial values $0,0.1,1$.
The zero-coefficient model is also evaluated explicitly. Optimization uses
L-BFGS-B with at most 500 iterations per start and requires at least one
converged solution.
Here $u$ is the training-weighted mean squared residual divided by two,
floored at $10^{-6}\,\mathrm{m}^2$, and $s_A$ is the training-weighted half
trace of $A$. Zero coefficient preserves H-Q exactly, without another floor.
This comparison tests the additional value of the anchored term within this
model family. Even a gain over H-QI would not distinguish contributions
from the observation-dependent magnitude and direction of $A$.

An outer leave-one-location-out evaluation covers all 300 calibration queries
at 24 locations: 262 positions and 38 failures. Each fold excludes its location
from all fitting, normalization and inner regularization selection. H-Q,
H-QA, H-QI and F-GA use the remaining samples from the common
calibration subset (248 in total). The F-C fallback uses all successful
outer-training samples. Each fold refits F-G, performs H-Q's inner
leave-one-location-out selection, and refits F-GA and both additions.
Missing G uses F-C;
missing A in H-QA uses H-Q; H-QI adds its term only when G is available.
F-GA uses F-C when either G or A is unavailable. Positions and failure states
are unchanged.

For validation and inPlace, F-GA and H-Q retain their earlier full-calibration fits,
and only the additive coefficients are fitted on the same 248 samples.
The coefficients are $c_A=0.788746$ and $c_I=0.002861$. Models and predictions
were fixed before either evaluation score was computed. NLL, coverage and area use successful outputs;
NR25 includes every eligible query and assigns failures risk one.

\begin{table*}[t]
\centering\small
\caption{Additive-model comparison using equal-location means. Source OOF uses
nested location holdout within the calibration data; validation and inPlace
reuse the earlier evaluation sets.
All cohorts belong to the same domain.}
\label{tab:conditional-anchor}
\begin{tabular}{@{}llrrrr@{}}
\toprule
Cohort & Model & NLL (nat) & PC90 (\%) & EA90 (m$^2$) & NR25 \\
\midrule
Source OOF & H-Q & 0.8936 & 90.22 & 8.775 & 0.138470 \\
Source OOF & H-QA & 0.2544 & 94.20 & 13.735 & 0.135774 \\
Source OOF & H-QI & 0.8603 & 91.21 & 8.825 & 0.138469 \\
Source OOF & F-GA & 0.1427 & 90.76 & 17.020 & 0.135331 \\
Val & H-Q & 0.2722 & 93.14 & 6.996 & 0.143556 \\
Val & H-QA & 0.2166 & 95.33 & 10.431 & 0.141186 \\
Val & H-QI & 0.1898 & 94.53 & 7.046 & 0.143556 \\
Val & F-GA & 0.1712 & 92.84 & 7.126 & 0.143292 \\
InPlace & H-Q & -0.4146 & 88.76 & 2.776 & 0.039049 \\
InPlace & H-QA & -0.4522 & 91.47 & 3.432 & 0.038361 \\
InPlace & H-QI & -0.5616 & 95.63 & 2.825 & 0.039049 \\
InPlace & F-GA & -0.6941 & 98.21 & 1.858 & 0.039224 \\
\bottomrule
\end{tabular}
\end{table*}

Descriptive intervals use 10,000 paired-location bootstrap resamples with seed 20261009.
They condition on fitted predictions and do not rerun model selection, so they
do not quantify uncertainty of the full development procedure.
H-QA has lower mean NR25 than H-Q in each cohort, but gains are concentrated in a few
locations and the source OOF and validation intervals include zero.

For source OOF predictions, H-QA minus H-Q has NLL difference $-0.6392$ with interval
$[-1.8621,0.0182]$ and NR25 difference $-0.002696$ with
$[-0.008064,0.000057]$. Classifying differences by their exact sign,
source NR25 has 3 improved locations, 0 ties and 21 regressed locations;
all 24 anchor coefficients
are positive. Excluding L12 changes the differences to
$-0.00007796$ for NR25 and $-0.06898$ for NLL; the median positive
NR25 regression is $1.49\times10^{-6}$. The main summaries include every location.

Validation NR25 difference is $-0.002370$ with $[-0.006894,0.000063]$,
with 5 improved locations, 0 ties and 19 regressed locations (median positive
regression $1.70\times10^{-5}$).

InPlace H-QA minus H-Q NR25 is $-0.000688$, with interval
$[-0.002064,0.000000043]$ across three previously evaluated locations; against
H-QI the interval ends just below zero ($-1.6\times10^{-8}$).

H-QA minus H-QI NR25 intervals also cross zero for source OOF and validation.
The complete-feature comparison uses successful outputs with both G and A:
248 source OOF, 585 validation and 132 inPlace outputs. Source OOF and
validation intervals again cross zero, leaving the incremental benefit
uncertain on this common subset.
H-QI has lower NLL than H-QA in both evaluation sets. Larger H-QA
regions and increased validation overcoverage accompany the mean-risk gain.
These comparisons do not establish equivalence between models.

\section{Discussion and Conclusion}
\subsection{Motivation for Base Anchoring}
The initial development evaluation produces 256 positions from 258 queries
at locations absent from the calibration set.
F-GD improves NLL from
2.0173 to $-0.2596$ relative to F-C, yet NR25 increases from 0.01379 to
0.01495, so improved predictive density did not translate into lower update
risk. Of twelve successful queries with increased risk, six have base errors
of only 0.065--0.669 m and account for 97.17\% of the summed positive risk
increments, rather than the net change. In these cases, increased uncertainty
reduces the weight of accurate observations, leaving more prior error.
Other regressions instead increase the observation-error contribution.
This analysis motivated F-GA, which was then evaluated on separate held-out queries.

Anchoring algebraically retains base offset as well as subset spread, but
the paired intervals do not establish a robust incremental advantage for
anchoring or selection alone.

\subsection{Ranking and Calibration}
\label{sec:anchor-ranking-scale-supplement}
An earlier within-location development analysis illustrates the distinction between
ranking and calibration. It uses the 496-record
\texttt{test\_outPlace} set at three locations and a different model,
F-GD-Config, calibrated by a single shared affine map of the configuration
disagreement. Its calibration and evaluation frames come from the same three
locations. Two evaluator defects were repaired after the labels had been
inspected, so the results are treated as a within-location development diagnostic.
Of 496 attempts, 493 produce positions. The split comprises 246
successful calibration and 247 successful evaluation records, assigned by sorted
query identity within location rather than a random half. Three failures were
not assigned to either uncertainty subset; availability is 493/496, while
the following uncertainty metrics use the fixed 247 evaluation outputs.
An update-risk score that includes every query is therefore unavailable for this cohort.

Mean absolute sparsification error (MASE) is computed from ranking and a
label-derived radius. The Gaussian models F-I, F-C and F-GD-Config rank
queries by the scalar score $s_i=\sqrt{\operatorname{tr}R_i/2}$.
At each $p=k/100$, $k=1,\ldots,100$, we retain the
$\max(1,\operatorname{round}(pn))$ smallest scores within a location.
The diagnostic radius is the largest residual norm in that subset, and
$q(p)$ is the fraction of all $n$ evaluation errors within that radius.
MASE averages $\mathbb E_{\rm ties}|q(p)-p|$ over the 100 levels, averaging
uniform tie orderings before summing absolute deviations. Labels construct
this diagnostic circle rather than a predictive region. MASE summarizes the
discrete grid, with no general discretization-error bound.

F-GD-Config improves MASE at all three locations, lowering the equal-location
mean from $0.4200$ to $0.3207$, while coverage deteriorates.
Here the coverage error is
$G^{-1}\sum_g|\mathrm{PC}_{0.90,g}-0.90|$, not the absolute error of
the macro coverage. At nominal $0.90$ the full-matrix control covers
$0.8530$ and F-GD-Config covers $0.8256$, so the coverage error rises from
$0.0946$ to $0.1506$, the largest of the three Gaussian models.
At the smallest location, with sixteen evaluation frames, F-GD-Config
improves MASE from $0.3833$ to $0.3431$ while worsening NLL from $11.0695$ to
$11.5352$ and coverage from $0.6875$ to $0.5625$. Thus, better ranking
coexists with worse probabilistic calibration in the same model and cohort.

The fitted scale map assigns similar predicted scales to locations with
markedly different error magnitudes. The shared map is $\sigma=3.302+3.687\,r_D$, with
$r_D=\sqrt{\operatorname{tr}D}$, under one
equal-location-weighted fit. Within-location rank correlation between
disagreement and error is $+0.408$, $+0.809$ and $+0.376$, while the median
predicted scale is $3.61$, $3.86$ and $4.06$~m against median errors of
$0.18$, $0.41$ and $9.88$~m. The ratio of median error to median scale is
therefore $0.05$, $0.11$ and $2.43$ across the three locations, spanning a
factor of $49$. Positive rank correlations coexist with a substantial
mismatch between the shared predicted scale and the error magnitude at each
location. This mismatch offers a plausible, though not independently verified,
explanation for the coverage results. The finding concerns this fitted scale
map; other mappings could extract different magnitude information from
configuration disagreement.

This divergence motivates evaluating predictive quality alongside update
risk at fixed positions. Its magnitude varies across the three locations:
$81.8\%$ of the MASE gain comes
from one location contributing 21 of 247 evaluation frames. In a separate
development run of the 300-query cohort, 263 positions were produced; 262 had
valid configuration disagreement at 24 locations. The median within-location
Spearman correlation between $r_D$ and horizontal reference-error magnitude
was $+0.003$. The associations differ between these development cohorts;
the available evidence does not explain that difference. The
label-derived MASE diagnostic alone is therefore insufficient to assess
scale or directional calibration.

\subsection{Limitations and Generalization}
In all experiments, the query imagery and maps share a reconstruction process.
The main validation and
inPlace comparisons cover held-out queries at calibration locations;
location-held-out analyses belong to method development.
Reference precision is unquantified, map errors can be correlated, and most
configurations can agree on an incorrect position. Fitted scales calibrate
reference residuals rather than identify physical DSM noise.

The measured gains concern weighting unchanged positions under an artificial
independent prior, not improved localization accuracy or validated navigation
performance. Transfer requires independently produced maps; global retrieval
and temporal evaluation require retrieval inputs and ordered flight sequences,
respectively. The incremental value beyond H-Q needs assessment at locations
not used during method development.

\subsection{Conclusion}
The benchmark links availability, error structure and predictive uncertainty
to the consequences of weighting position observations. In the case study,
base-anchored calibration improves predictive density and reduces the
prespecified equal-location update risk relative to the constant full-matrix
baseline by 7.70\% and 5.46\% on two sets of
held-out queries at represented locations, using reference residuals where
positions are available and leaving the position estimates unchanged. The update uses a controlled
independent prior. Exploratory H-Q and
conditional-addition comparisons leave the incremental update-risk benefit
of configuration disagreement unresolved. Independent-location evaluation
and ordered flight sequences are needed to assess transfer and navigation use.

\bibliographystyle{IEEEtran}
\bibliography{references}

\begin{thebibliography}{10}
\providecommand{\url}[1]{#1}
\csname url@samestyle\endcsname
\providecommand{\newblock}{\relax}
\providecommand{\bibinfo}[2]{#2}
\providecommand{\BIBentrySTDinterwordspacing}{\spaceskip=0pt\relax}
\providecommand{\BIBentryALTinterwordstretchfactor}{4}
\providecommand{\BIBentryALTinterwordspacing}{\spaceskip=\fontdimen2\font plus
\BIBentryALTinterwordstretchfactor\fontdimen3\font minus
  \fontdimen4\font\relax}
\providecommand{\BIBforeignlanguage}[2]{{%
\expandafter\ifx\csname l@#1\endcsname\relax
\typeout{** WARNING: IEEEtran.bst: No hyphenation pattern has been}%
\typeout{** loaded for the language `#1'. Using the pattern for}%
\typeout{** the default language instead.}%
\else
\language=\csname l@#1\endcsname
\fi
#2}}
\providecommand{\BIBdecl}{\relax}
\BIBdecl

\bibitem{vakhitov2021uncertainty}
\BIBentryALTinterwordspacing
A.~Vakhitov, L.~Ferraz, A.~Agudo, and F.~Moreno-Noguer, ``Uncertainty-aware
  camera pose estimation from points and lines,'' in \emph{Proceedings of the
  IEEE/CVF Conference on Computer Vision and Pattern Recognition}, 2021, pp.
  4659--4668. [Online]. Available:
  \url{https://openaccess.thecvf.com/content/CVPR2021/papers/Vakhitov_Uncertainty-Aware_Camera_Pose_Estimation_From_Points_and_Lines_CVPR_2021_paper.pdf}
\BIBentrySTDinterwordspacing

\bibitem{levi2020evaluating}
\BIBentryALTinterwordspacing
D.~Levi, L.~Gispan, N.~Giladi, and E.~Fetaya, ``Evaluating and calibrating
  uncertainty prediction in regression tasks,'' 2020, version 3. [Online].
  Available: \url{https://arxiv.org/abs/1905.11659v3}
\BIBentrySTDinterwordspacing

\bibitem{kuppers2022parametric}
\BIBentryALTinterwordspacing
F.~K{\"u}ppers, J.~Schneider, and A.~Haselhoff, ``Parametric and multivariate
  uncertainty calibration for regression and object detection,'' 2022, version
  2; SAIAD workshop at ECCV. [Online]. Available:
  \url{https://arxiv.org/abs/2207.01242v2}
\BIBentrySTDinterwordspacing

\bibitem{detone2018superpoint}
\BIBentryALTinterwordspacing
D.~DeTone, T.~Malisiewicz, and A.~Rabinovich, ``{SuperPoint}: Self-supervised
  interest point detection and description,'' in \emph{Proceedings of the IEEE
  Conference on Computer Vision and Pattern Recognition Workshops}, 2018, pp.
  224--236. [Online]. Available:
  \url{https://openaccess.thecvf.com/content_cvpr_2018_workshops/w9/html/DeTone_SuperPoint_Self-Supervised_Interest_CVPR_2018_paper.html}
\BIBentrySTDinterwordspacing

\bibitem{lindenberger2023lightglue}
\BIBentryALTinterwordspacing
P.~Lindenberger, P.-E. Sarlin, and M.~Pollefeys, ``{LightGlue}: Local feature
  matching at light speed,'' in \emph{Proceedings of the IEEE/CVF International
  Conference on Computer Vision}, 2023, pp. 17\,627--17\,638. [Online].
  Available:
  \url{https://openaccess.thecvf.com/content/ICCV2023/html/Lindenberger_LightGlue_Local_Feature_Matching_at_Light_Speed_ICCV_2023_paper.html}
\BIBentrySTDinterwordspacing

\bibitem{edstedt2024roma}
\BIBentryALTinterwordspacing
J.~Edstedt, Q.~Sun, G.~B{\"o}kman, M.~Wadenb{\"a}ck, and M.~Felsberg, ``{RoMa}:
  Robust dense feature matching,'' in \emph{Proceedings of the IEEE/CVF
  Conference on Computer Vision and Pattern Recognition}, 2024. [Online].
  Available:
  \url{https://openaccess.thecvf.com/content/CVPR2024/papers/Edstedt_RoMa_Robust_Dense_Feature_Matching_CVPR_2024_paper.pdf}
\BIBentrySTDinterwordspacing

\bibitem{shen2024gim}
\BIBentryALTinterwordspacing
X.~Shen, Z.~Cai, W.~Yin, M.~M{\"u}ller, Z.~Li, K.~Wang, X.~Chen, and C.~Wang,
  ``{GIM}: Learning generalizable image matcher from internet videos,'' in
  \emph{The Twelfth International Conference on Learning Representations},
  2024. [Online]. Available:
  \url{https://proceedings.iclr.cc/paper_files/paper/2024/hash/38830ee2c33d56682385a00c8e7e9b85-Abstract-Conference.html}
\BIBentrySTDinterwordspacing

\bibitem{dhaouadi2025ortholoc}
\BIBentryALTinterwordspacing
O.~Dhaouadi, R.~Marin, J.~M. Meier, J.~Kaiser, and D.~Cremers, ``{OrthoLoC}:
  {UAV} {6-DoF} localization and calibration using orthographic geodata,'' in
  \emph{The Thirty-ninth Annual Conference on Neural Information Processing
  Systems Datasets and Benchmarks Track}, 2025. [Online]. Available:
  \url{https://deepscenario.github.io/OrthoLoC/}
\BIBentrySTDinterwordspacing

\bibitem{ye2026anyvisloc}
\BIBentryALTinterwordspacing
Y.~Ye, X.~Teng, S.~Chen, L.~Liu, K.~Wang, X.~Song, and Z.~Li, ``Exploring the
  best way for {UAV} visual localization under low-altitude multi-view
  observation condition: A benchmark,'' 2026, version 2. [Online]. Available:
  \url{https://arxiv.org/abs/2503.10692v2}
\BIBentrySTDinterwordspacing

\bibitem{cerescovariance}
\BIBentryALTinterwordspacing
{Ceres Solver Developers}, ``Covariance estimation,'' Ceres Solver
  documentation, accessed September 26, 2026. [Online]. Available:
  \url{https://ceres-solver.readthedocs.io/latest/nnls_covariance.html}
\BIBentrySTDinterwordspacing

\bibitem{kang2026ricloc}
\BIBentryALTinterwordspacing
W.~Kang, J.~Kim, J.~Lee, and T.-W. Kim, ``Reference-induced consensus for
  selective posed-reference visual localization,'' 2026, version 1. [Online].
  Available: \url{https://arxiv.org/abs/2607.04722v1}
\BIBentrySTDinterwordspacing

\bibitem{hubert2018mcd}
\BIBentryALTinterwordspacing
M.~Hubert, M.~Debruyne, and P.~J. Rousseeuw, ``Minimum covariance determinant
  and extensions,'' \emph{WIREs Computational Statistics}, vol.~10, no.~3, p.
  e1421, 2018. [Online]. Available: \url{https://doi.org/10.1002/wics.1421}
\BIBentrySTDinterwordspacing

\bibitem{elmhamdi2020mda}
\BIBentryALTinterwordspacing
E.-M. El-Mhamdi, R.~Guerraoui, A.~Guirguis, L.~N. Hoang, and S.~Rouault,
  ``Genuinely distributed byzantine machine learning,'' in \emph{Proceedings of
  the 39th Symposium on Principles of Distributed Computing}, 2020, pp.
  355--364. [Online]. Available: \url{https://doi.org/10.1145/3382734.3405695}
\BIBentrySTDinterwordspacing

\bibitem{gneiting2007proper}
\BIBentryALTinterwordspacing
T.~Gneiting and A.~E. Raftery, ``Strictly proper scoring rules, prediction, and
  estimation,'' \emph{Journal of the American Statistical Association}, vol.
  102, no. 477, pp. 359--378, 2007. [Online]. Available:
  \url{https://sites.stat.washington.edu/people/raftery/Research/PDF/Gneiting2007jasa.pdf}
\BIBentrySTDinterwordspacing

\end{thebibliography}
\end{document}